\documentclass{ieeeaccess}
\usepackage{cite}
\usepackage{amsmath,amssymb,amsfonts}
\usepackage{algorithmic}
\usepackage{graphicx}
\usepackage{textcomp}
\usepackage{subcaption}
\usepackage{hyperref}
\usepackage[capitalize]{cleveref}
\usepackage{booktabs}
\usepackage{makecell}
\usepackage{pifont}
\usepackage[table]{xcolor}
\usepackage{multirow}

\def\BibTeX{{\rm B\kern-.05em{\sc i\kern-.025em b}\kern-.08em
    T\kern-.1667em\lower.7ex\hbox{E}\kern-.125emX}}
\begin{document}
\history{Date of publication xxxx 00, 0000, date of current version xxxx 00, 0000.}
\doi{10.1109/ACCESS.2017.DOI}

\newcommand{\orcidlink}[1]{\href{https://orcid.org/#1}{\textcolor[HTML]{A6CE39}{\textsuperscript{\textregistered}}}}

%
\title{MDS-DETR: DETR with Masked Duplicate Suppressor}

\author{\uppercase{Chanho Lee}\orcidlink{0009-0002-9762-6386}\authorrefmark{1}$^*$, \uppercase{Seunghee Koh}\orcidlink{0009-0006-8662-0834}\authorrefmark{2},
\uppercase{Yunho Jeon}\orcidlink{0000-0001-8043-480X}\authorrefmark{3}, and \uppercase{Junmo Kim}\orcidlink{0000-0002-7174-7932}\authorrefmark{2}\IEEEmembership{Member, IEEE}.}

\address[1]{Samsung Research, Seoul 06765, South Korea (email: yiwan8833@gmail.com)}
\address[2]{Korea Advanced Institute of Science and Technology (KAIST), Daejeon 34141, South Korea}
\address[3]{Department of Artificial Intelligence Software, Hanbat National University, Daejeon 34158, South Korea}
\tfootnote{*This work was conducted while Chanho Lee was with the Korea Advanced Institute of Science and Technology (KAIST). \newline
This work was partly supported by Center for Applied Research in Artificial Intelligence (CARAI) grant funded by Defense Acquisition Program Administration (DAPA) and Agency for Defense Development (ADD) (UD230017TD) and a grant of the Korea Health Technology R\&D Project through the Korea Health Industry Development Institute (KHIDI), funded by the Ministry of Health \& Welfare, Republic of Korea (grant number: RS-2022-KH129703). }


\markboth
{Author \headeretal: Preparation of Papers for IEEE TRANSACTIONS and JOURNALS}
{Author \headeretal: Preparation of Papers for IEEE TRANSACTIONS and JOURNALS}

\corresp{Corresponding author: Junmo Kim (e-mail: junmo.kim@kaist.ac.kr), Yunho Jeon (yhjeon@hanbat.ac.kr)}

\begin{abstract}
The DEtection TRansformer (DETR) is a powerful end-to-end object detector, yet its one-to-one matching strategy suffers from slow convergence and low recall. A common approach to address this issue is to use one-to-many label assignment to provide more positive samples. However, existing methods that use one-to-many matching as an auxiliary objective lead to increased training costs, with their auxiliary decoders discarded during inference.
To address this limitation, we propose MDS-DETR, which leverages both one-to-one and one-to-many supervision within a single decoder. Specifically, we introduce a Masked Duplicate Suppressor (MDS) that injects asymmetry into self-attention via confidence-based causal masking. MDS filters out the duplicates generated by the one-to-many supervised layer, enables explainable, duplicate-free predictions in a fully end-to-end framework.
MDS-DETR outperforms existing one-to-many DETR variants such as MS-DETR, MR.DETR and Relation-DETR, without relying on any additional queries or auxiliary decoders. Under a 12-epoch training schedule on MS COCO with a ResNet-50 backbone, MDS-DETR achieves a +2.8 mAP improvement over Deformable-DETR with only a 5\% increase in training time, and outperforms the state-of-the-art MR.DETR by +0.3 mAP while being even 20\% faster in training. Our code and models are available at \href{https://github.com/dcholee/mds-detr}{https://github.com/DChoLee/MDS-DETR}.
\end{abstract}


\titlepgskip=-15pt

\maketitle

\section{Introduction}
\label{sec:introduction}
DEtection TRansformer (DETR)~\cite{detr} is a fully end-to-end object detector that utilizes Hungarian one-to-one matching. It eliminates the need for hand-crafted components such as anchor boxes and non-maximum suppression (NMS), enabling duplicate-free predictions. However, one-to-one matching suffers from slow convergence due to the limited number of positive queries. Hybrid-DETR~\cite{jia2023detrs} addresses this problem by training an auxiliary weight-shared decoder with one-to-many supervision to enrich positive matches and accelerate convergence. Recently, most DETRs with one-to-many supervision~\cite{jia2023detrs,chen2023group,zong2023detrs,pu2024rank,hou2024relation,zhang2024mr} have boosted their performance by incorporating auxiliary decoders during training.

Nevertheless, this strategy substantially increases training cost, as it requires additional queries, supervisions, and corresponding losses. For instance, MR.DETR~\cite{zhang2024mr} trains one main decoder with one-to-one supervision alongside two auxiliary decoders with one-to-many supervision, leading up to 30\% longer training time compared to Deformable-DETR~\cite{zhu2020deformable}. Although these auxiliary decoders are discarded during inference, they even outperform the main decoder when post-processed with NMS. In this paper, we argue that training multiple auxiliary decoders up to state-of-the-art levels merely to support a main decoder is inefficient and suboptimal. Through our experiments, one-to-many supervision can be effectively integrated into training a single decoder, while even achieving better performance. 

As classical object detectors produce highly dense box predictions that are subsequently filtered by NMS to remove duplicates, we aim to design our DETR decoder such that earlier layers are trained with one-to-many supervision to output dense object queries, and subsequent layers are trained with one-to-one supervision to produce duplicate-free predictions. Thus, the one-to-one supervised layers are expected to identify true positive queries and suppress false positive queries as duplicate, similarly to NMS. However, as shown in \cite{jia2023detrs}, this so-called hybrid layer scheme yields degraded performance. This failure is not from the hybrid scheme itself, but from the inherent symmetry of the DETR decoder.

Duplicate suppression inherently requires high asymmetry: even when multiple candidate boxes have exactly the same bounding boxes and classification scores, only one should be preserved while the others must be suppressed. In contrast, if two object queries are identical, their predictions obtained by forwarding through the DETR decoder will also be perfectly identical, because DETR defines object queries as a set rather than a sequence and the decoder’s self-attention is permutation-equivariant. Therefore, a conventional decoder layer would struggle to function as a duplicate suppressor.

Several prior works have attempted to address this inherent symmetry of DETR decoders. Rank-DETR~\cite{pu2024rank} introduces confidence-based rank embeddings to impose asymmetry. While this is a straightforward approach, it often leads the decoder to overfit to the absolute rank itself, rather than leveraging meaningful interactions through self-attention.
The mixed query selection strategy in DINO~\cite{zhang2022dino} can also be interpreted as a proposal confidence-based embedding for the initial query. However, its primary role is to facilitate query denoising as a context embedding, rather than acting as a rank embedding.
Ease-DETR~\cite{gao2024ease} takes a different approach by explicitly encoding both relative rank and pairwise IoU as attention weights. While IoU-based weights encourage interactions among highly overlapped queries, they lack the capacity to model fine-grained spatial relationships between queries.
 
\begin{figure*}[t]
  \centering
    \begin{subfigure}[b]{0.42\linewidth}
      \includegraphics[width=\linewidth]{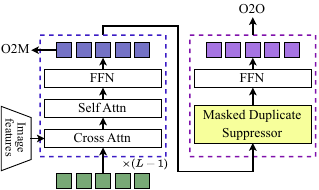}
      \caption{Decoder structure of our MDS-DETR}
      \label{fig:left}
    \end{subfigure}
  \hspace{5pt}
    \begin{subfigure}[b]{0.42\linewidth}
      \includegraphics[width=\linewidth]{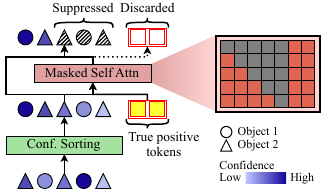}
      \caption{Masked Duplicate Suppressor}
      \label{fig:right}
    \end{subfigure}
  \caption{(a) The MDS-DETR decoder is composed of two parts: a sequence of O2M(one-to-many) layers followed by a single O2O(one-to-one) layer. The one-to-many layers iteratively refine object queries using the image features extracted by the backbone and encoder. While the one-to-many layers produce a dense set of high-qualified redundant candidate queries, the final layer is trained with one-to-one matching, acts as a learned duplicate suppressor. (b) Confidence-based sorting and attention masking ensure that each query can only attend to higher-confidence queries. When duplicate queries are successfully suppressed by higher-confidence queries, true positive queries may end up with inappropriate attention scopes, or all their attentions are blocked. To address this, we introduce a true positive token into the key and value, providing an “attention sink” that guarantees a true positive signal. After the masked self-attention, we apply an inverse sort operation to restore the original order of the queries. For simplicity, this step is omitted in the figure.}
  \label{fig:three_part_layout}
\end{figure*}

To handle this issue, we introduce a novel Masked Duplicate Suppressor (MDS) designed as an asymmetric block for the one-to-one supervised decoder layer. As illustrated in \cref{fig:three_part_layout}, MDS consists of confidence-based sorting and masked self-attention, based on the assumption that the queries obtained from one-to-many supervised layers are reliable and their confidences are well-calibrated. While NMS operates as an iterative post-processing method in which \textit{higher-confidence boxes suppress lower-confidence ones} using handcrafted IoU thresholds, our masked self-attention instead learns \textit{how lower-confidence queries are suppressed by higher-confidence queries} in parallel,  without relying on any handcrafted hyperparameters.

Our MDS-DETR is clearly distinguished from previous DETRs that focus on the crucial role of self-attention as a duplicate-free prediction. Prior approaches enforced self-attention to act as a duplicate suppressor by introducing auxiliary decoders or learnable query embeddings, which eventually leads its behavior to be intractable. In contrast, our novel masked self-attention and true positive tokens are explicitly designed for duplicate suppression, offering explainability. Through comprehensive experiments on the COCO 2017 benchmark~\cite{mscoco}, we demonstrate that MDS-DETR outperforms state-of-the-art DETRs such as MS-DETR~\cite{zhao2024ms}, MR.DETR~\cite{zhang2024mr}, and Relation-DETR~\cite{hou2024relation}. Furthermore, our method significantly reduces computational complexity, as it does not require any additional queries, losses, or any auxiliary decoders.

\section{Related Work}
\subsection{Detection transformers} The DETR framework introduced a fully end-to-end paradigm for object detection and successfully extended to a wide range of vision recognition tasks ~\cite{misra2021end,cheng2022masked,meinhardt2022trackformer,yu2022cmt,li2021pose}. Subsequent DETR-based variants have focused mainly on addressing the slow convergence of DETR and the limited training efficiency. Deformable DETR~\cite{zhu2020deformable} replaces global dense attention with a multi-scale deformable attention module that restricts attention to a sparse set of key features around reference points. DAB-DETR~\cite{liu2022dabdetr} proposes dynamic anchor boxes as object queries, allowing the model to refine both query content and position iteratively. DN-DETR~\cite{li2022dn} and DINO-DETR~\cite{zhang2022dino} further accelerate training and enhance both stability and supervision by introducing denoising-based query learning and mixed query selection strategies. While most existing DETR-based detectors focus on improving object query design~\cite{liu2022dabdetr,wang2022anchor} or enhancing image feature aggregation via cross-attention~\cite{gao2021fast,hou2024salience}, we focus on the ability of self-attention in the decoder to suppress duplicated candidates for one-to-one prediction.

\subsection{DETRs with one-to-many supervision} One-to-many supervision assigns multiple queries to each ground truth object for significantly accelerating convergence and improving recall in DETR-based models. Group-DETR~\cite{chen2023group} introduces a group-wise one-to-many assignment strategy to enhance positive supervision signals. Hybrid-DETR~\cite{jia2023detrs} proposes a weight-shared auxiliary decoder trained with one-to-many matching, referred to as a hybrid branch. This hybrid branch scheme has been widely adopted in subsequent one-to-many-based DETR variants such as DAC-DETR~\cite{hu2024dac}, Rank-DETR~\cite{pu2024rank}, Relation-DETR~\cite{hou2024relation}, and MR.DETR~\cite{zhang2024mr}. However, the use of auxiliary decoders inherently requires a large number of additional queries, which leads to a significant trade-off between training complexity and performance. In this paper, we question the necessity of auxiliary decoders and propose the Masked Duplicate Suppressor (MDS) as an effective alternative. MDS-DETR achieves performance that is competitive with, and in some cases superior to, existing heavily trained one-to-many-based DETRs, while introducing only negligible additional training cost compared to the Deformable-DETR baseline. 

\subsection{Rank-oriented detection transformers} In classical object detection frameworks, rank-oriented designs have primarily focused on constructing IoU-aware confidence measures through loss functions~\cite{li2020generalized,zhang2021varifocalnet} and architectural modifications~\cite{zhang2021varifocalnet,wu2020iou,feng2021tood}. In contrast, DETR not only suffers from such misalignment but also faces the challenge of suppressing duplicate predictions. To address this, several DETR variants have explored rank-oriented mechanisms. Rank-DETR~\cite{pu2024rank} introduces a learnable confidence-rank embedding with a rank-aware classification head. Align-DETR~\cite{cai2023align} proposes a rank-scaled loss function applied across one-to-many-matched queries, down-weighting the loss from lower-rank queries. Ease-DETR~\cite{gao2024ease} encourages queries with higher confidence ranks to remain dominant in subsequent decoder layers, while those with lower ranks are gradually suppressed. Unlike methods such as DETA~\cite{ouyang2022nms}, DDQ-DETR~\cite{ddqdetr}, and DE-DETR~\cite{wang2022towards} that explicitly incorporate NMS during training and inference, our MDS encodes confidence-rank causality directly into the attention mask. Motivated by the idea that higher-ranked queries correspond to true positives, we propose a confidence rank-based masked self-attention mechanism. For duplicate suppression, we restrict the direction of attention to flow only from lower- to higher-ranked queries, which can be interpreted as a form of parallelized NMS. Our rank-based masking design enables fully end-to-end learning without any additional post-processing or hand-crafted design.

\section{Preliminaries}
\subsection{DETR architecture.} A general DETR architecture consists of a backbone, an encoder, and a decoder. The input image is first processed by the backbone to extract multi-scale feature maps, which are then passed through the encoder to generate a global image embedding. Given an initial set of $n$ object queries \(Q^{(0)} = \{\mathbf{q}^{(0)}_1, \ldots, \mathbf{q}^{(0)}_{n} \}\), the decoder progressively refines the queries to produce the final set \(Q^{(L)} = \{ \mathbf{q}^{(L)}_1, \ldots, \mathbf{q}^{(L)}_{n} \}\). Every decoder layer has both a classification and a localization head, where each query \( \mathbf{q}^{(l)}_i \in Q^{(l)} \) is mapped to a classification score vector \( \mathbf{s}^{(l)}_i \in \mathbb{R}^C \), where \(C\) is the number of classes, and a bounding box prediction \( \mathbf{b}^{(l)}_i \in \mathbb{R}^4 \). The outputs \( S^{(l)} = \{ \mathbf{s}^{(l)}_1, \ldots, \mathbf{s}^{(l)}_{n} \} \) and \( B^{(l)} = \{ \mathbf{b}^{(l)}_1, \ldots, \mathbf{b}^{(l)}_{n} \} \) are supervised using either one-to-one or one-to-many matching strategies, depending on the training objective.
\subsection{One-to-one matching strategy.}
One-to-one matching assigns a single object query to each ground truth object.  
Given a ground truth set of $m$ objects denoted by $\bar{B} = \{\mathbf{\bar{b}}_1, \dots, \mathbf{\bar{b}}_{m} \}$ and $\bar{S} = \{\mathbf{\bar{s}}_1, \dots, \mathbf{\bar{s}}_{m} \}$ representing their bounding boxes and class labels, respectively, the one-to-one matcher finds a permutation function $\sigma(\cdot)$ via bipartite matching that minimizes the matching cost between predictions and ground truth. The unmatched predictions are then treated as background. Then, one-to-one supervision loss is defined as:
\begin{equation}
    \mathcal{L}_{\mathrm{one2one}} = \sum_{i=1}^{n} \left[ \mathcal{L}_{\mathrm{cls}}\left(\mathbf{s}_i,\, \mathbf{\bar{s}}_{\sigma(i)}\right) + \mathcal{L}_{\mathrm{bbox}}\left(\mathbf{b}_i,\, \mathbf{\bar{b}}_{\sigma(i)}\right) \right],
\end{equation}

where $\mathcal{L}_{\mathrm{cls}}$ is the classification loss, and $\mathcal{L}_{\mathrm{bbox}}$ consists of the L1 and GIoU losses for bounding box regression. The property of one-to-one matching that sends all unmatched queries to background encourages DETR to produce duplicate-free predictions. In our model, one-to-one supervision is applied only to the final decoder layer.

\subsection{One-to-many matching strategy.}
One-to-many matching allows multiple object queries to be assigned to a single ground truth object. Since most DETRs with one-to-many matching~\cite{zhang2024mr,zhao2024ms,hu2024dac} are built upon the matching strategy proposed in DETA~\cite{ouyang2022nms}, our method also adopts this matching formulation. Given a prediction $(\mathbf{s}_i, \mathbf{b}_i)$ from the $i$-th query and a ground truth $(\mathbf{\bar{s}}_j, \mathbf{\bar{b}}_j)$ for the $j$-th object, the matching score is defined as:
\begin{equation}
    \mathrm{MatchScore}_{ij} = \alpha \cdot \mathbf{s}_i^T  \mathbf{\bar{s}}_j + (1 - \alpha) \cdot \mathrm{IoU}(\mathbf{b}_i, \mathbf{\bar{b}}_j)
\end{equation}

where $\alpha \in [0,1]$ is a balancing weight between classification confidence and localization quality. The one-to-many matcher selects the top-$K$ queries for each ground truth object based on the match score, and discards low-quality matches with scores fall below a predefined threshold $\tau$. This swipe-out mechanism effectively improves the stability of one-to-many supervision, particularly during the early stages of training. In MDS-DETR, all decoder layers except the final one are trained with one-to-many supervision loss.

\section{DETR with Masked Duplicate Suppressor}
The MDS-DETR decoder is built upon a hierarchical matching strategy that integrates both one-to-many and one-to-one supervision within a single decoder. As illustrated in \cref{fig:left}, the earlier decoder layers are trained with one-to-many supervision to generate a high-quality set of candidate queries. Then, the last decoder layer, composed of Masked Duplicate Suppressor (MDS) and a feed-forward-network (FFN), is trained with one-to-one supervision. MDS selects true positives among these candidates and learns to suppress the remaining queries as duplicates. The one-to-many layers follow a Deformable DETR decoder where the order of cross-attention and self-attention is reversed. As reported in MS-DETR~\cite{zhao2024ms}, this rearrangement yields a marginal performance improvement of 0.1 mAP. To ensure that MDS focuses solely on duplicate suppression, we remove cross-attention and any further localization refinement for the one-to-one layer. This design ensures that the MDS layer solely contributes to the classification and suppression of duplicate queries. In the following section, we present the core components of MDS, including confidence-based sorting, the masked self-attention mechanism, and the true positive token. Then, we describes the formulation of the relative position bias and its integration into the masked self-attention for effective duplicate suppression.

\subsection{Masked duplicate suppressor}
In one-to-many matching, we refer to the set of $K$ queries assigned to a single ground truth object as candidate queries with the \textit{same-context}. When the one-to-many predictions are reliable, the true positive can be selected as the highest-confidence query among them. The key to differentiating this single true positive from the remaining $K-1$ duplicates lies in whether higher-confidence query with the \textit{same-context} exists or not. Based on this observation, our strict masking formulation enables duplicate suppression to be reframed as a task of detecting whether a \textit{same-context} query exists within the attention scope.

In this perspective, our Masked Duplicate Suppressor(MDS) consists of confidence-based sorting and masked self-attention. Let MDS is receiving input queries $Q^{(l-1)}$ with class probabilities $S^{(l-1)}$ and corresponding bounding boxes $B^{(l-1)}$ from the previous one-to-many supervised layer. Since the residual connection preserves the original input, duplicate queries would receive suppressive updates from the masked self-attention, guiding them toward background by canceling their residual signal. By contrast, true positive queries retain their original inputs and receive minimal updates from the masked self-attention, allowing them to remain unsuppressed. To construct this design, we first sort the query set $Q^{(l-1)}$ in descending order based on the maximum class probability from each query’s classification score, denoted as $S^{(l-1)}{\text{max}}$, to obtain the sorted set $\hat{\mathcal{Q}}^{(l-1)}$. Then, we define the self-attention mask as:
\begin{equation}
    \hat{M}_{ij} =
    \begin{cases}
    1, & \text{if } i > j \\
    0, & \text{otherwise},
    \end{cases}
\end{equation}
where the mask $\hat{M}_{ij}$ blocks query $\mathbf{\hat{q}}_i$ from attending to all lower-confidence queries $\mathbf{\hat{q}}_j$ as $i \leq j$. Then duplicate queries are only allowed to attend to higher-confidence ones, naturally directing their attention toward the overlapping true positive query. This interaction enables the masked self-attention to suppress the original input retained by the residual connection, leading to duplicate suppression behavior.

However, true positive queries may not find any \textit{same-context} queries within their attention scope, potentially causing them to attend to unrelated, different-context queries. To address this, we introduce a \textbf{true positive token} that absorbs attention from true positives when no valid suppression target exists. This token acts as an attention sink and guides true positives to maintain their predictions without being distorted by irrelevant attention. This design is also consistent with our earlier interpretation of residual connections that preserve one-to-many predictions.

\subsection{Symmetric relative position bias}
\label{sec:relative_position_bias}
Most DETR-based methods incorporate absolute position embeddings to provide accurate localization and classification of objects within an image, but the relative position between queries plays a more critical role for duplicate suppression. Following prior works~\cite{hu2018relation, hou2024relation}, we define the relative position encoding between two queries as:
{\small
\begin{equation}
\mathbf{e}_{ij}= \left[
\log\left(\frac{|\Delta x|}{w_i}+1\right),
\log\left(\frac{|\Delta y|}{h_i}+1\right),
\log\left(\frac{w_i}{w_j}\right),
\log\left(\frac{h_i}{h_j}\right)
\right]
\end{equation}
}

\definecolor{Gray}{gray}{0.95} 
\begin{table*}[t]
  \centering
  \caption{Comparison with state-of-the-art methods on COCO \texttt{val2017} with a
ResNet-50 backbone. We verified whether each method employs any auxiliary decoder that imposes additional training cost. DINO's auxiliary denoising queries are not considered, as their computational overhead is negligible. The \textbf{best} result is shown in bold.}
  \label{tab:exp coco2017}
  \resizebox{\textwidth}{!}{
  \begin{tabular}{@{}l|cc|c|ccccccc@{}}
    \toprule
    Method                                      & Queries  & Epochs \hspace{2pt} & \hspace{3pt} \makecell{Auxiliary\\Decoder} \hspace{3pt} & \hspace{5pt} mAP \hspace{5pt}    & AP$_{50}$         & AP$_{75} $        & AP$_S$        & AP$_M$        & AP$_L$        \\ \midrule
    Deformable-DETR++         & 300 & 12 & \ding{55}    & 47.0          & 65.3          & 51.0          & 30.1          & 50.5          & 60.7          \\
    Hybrid-DETR   & 300 & 12  & \checkmark    & 48.7          & 66.4          & 52.9          & 31.2          & 51.5          & 63.5          \\
    MS-DETR                    & 300 & 12  & \ding{55}   & 48.8          & 66.2          & 53.2         & 31.5          & 52.3          & 63.7          \\
    MR.DETR                  & 300 & 12   & \checkmark  &  
    49.5          & 67.0          & 53.7         & 32.1          & 52.5          & 64.7          \\
    \rowcolor{Gray}
    \textbf{MDS-DETR}                   & 300 & 12  & \ding{55}   &  
    \textbf{49.8}         & \textbf{67.1}          & \textbf{54.2}         & \textbf{32.2}          & \textbf{53.6}          & \textbf{65.2}          \\
    

    
   
    
    \midrule
    Deformable-DETR++        & 900 & 12  & \ding{55}   & 47.6          & 65.8          & 51.8          & 31.2        & 50.6          & 62.6          \\
    Salience-DETR        & 900 & 12  & \ding{55}   & 49.2          & 67.1          & 53.8          & 32.7          & 53.0          & 63.1          \\
    
    DINO               & 900 & 12  & \ding{55}   & 49.9          & 67.4          & 54.5          & 33.9          & 53.5          & 63.8          \\
    MS-DETR                    & 900 & 12  & \ding{55}   & 50.0          & 67.3          & 54.4         & 31.6          & 53.2          & 64.0          \\
    DAC-DETR                   & 900 & 12   & \checkmark  & 50.0          & 67.6          & 54.7          &  32.9             &  53.1             & 64.2              \\
    Rank-DETR                 & 900 & 12   & \checkmark  & 50.4          & 67.9          & 55.2          & 33.6          & 53.8          & 64.2          \\
    MR.DETR                  & 900 & 12  & \checkmark   &  
    50.7          & 68.2          & 55.4         & 33.6          & 54.3          & 64.6          \\
    \rowcolor{Gray}
    \textbf{MDS-DETR}                   & 900 & 12  & \ding{55}   &  
    51.1          & 68.8          & 55.5         & \textbf{34.0}          & \textbf{55.0}          & 66.0          \\
    \rowcolor{Gray}
    \textbf{Hybrid-MDS-DETR} \hspace{15pt}                  & 900 & 12   & \checkmark  &  
    \textbf{51.2}        & \textbf{68.9}         & \textbf{55.8}         & \textbf{34.0}          & 54.8          & \textbf{66.2}          \\
    \midrule
    Deformable-DETR++       & 900 & 24   & \ding{55}  & 49.8          & 67.0          & 54.2          & 31.4        & 52.8          & 64.1          \\
    DINO                    & 900 & 24  & \ding{55}   & 50.4          & 68.3          & 54.8          & 33.3          & 53.7          & 64.8          \\
    Salience-DETR         & 900 & 24   & \ding{55}  & 51.2          & 68.9          & 55.7          & 33.9          & 55.5          & 65.6          \\
    DAC-DETR                   & 900 & 24  & \checkmark   & 51.2          & 68.9          & 56.0          &  34.0             &  54.6             & 65.4              \\
    Align-DETR            & 900 & 24   & \ding{55}  & 51.3          & 68.2          & 56.1          & 35.5          & 55.1          & 65.6          \\
    MR.DETR                 & 900 & 24   & \checkmark  &  
    51.4          & 69.0          & 56.2         & 34.9          & 54.8          & 66.0          \\
    MS-DINO                    & 900 & 24   & \ding{55}  & 51.7          & 68.7          & 56.5         & 34.0          & 55.4          & 65.5          \\
    MR.DINO                & 900 & 24  & \checkmark   &  
    51.7          & 69.2          & 56.4         & 34.1          & 55.1          & 65.8          \\
    Relation-DINO                              & 900 & 24  & \checkmark   & 52.1 & \textbf{69.7} & 56.6          & \textbf{36.1} & \textbf{56.0} & 66.5 \\
    \rowcolor{Gray}
    \textbf{MDS-DETR}                   & 900 & 24  & \ding{55}   &  
    \textbf{52.3}         & \textbf{69.7}          & \textbf{56.9}         & 35.5          & \textbf{56.0}          & \textbf{67.0}          \\
    \bottomrule
  \end{tabular}
  }
\end{table*}

\definecolor{Gray}{gray}{0.95} 
\begin{table}[h]
  \caption{Comparison with state-of-the-art methods on COCO \texttt{val2017} with Swin-L~\cite{liu2021swin} backbone.}
  \label{tab:swin}
  \centering
  \resizebox{\columnwidth}{!}{
  \begin{tabular}{@{}lccccccccc@{}}
    \toprule
    \multicolumn{1}{c}{Method}   & mAP      & AP$_{50}$         & AP$_{75} $        & AP$_S$        & AP$_M$        & AP$_L$        \\ \midrule
    
    DINO \cite{zhang2022dino}                         & 56.8          & 75.4          & 62.3          & 41.1          & 60.6          & 73.5          \\
    Salience-DETR \cite{hou2024salience}              & 56.5          & 75.0          & 61.5          & 40.2          & 61.2          & 72.8          \\
    DAC-DETR \cite{hu2024dac}                         & 57.3          & 75.7          & 62.7          &  40.1             &  61.5             & 74.4              \\
    Stable-DINO \cite{liu2023detection}                    & 57.7          & 75.7         & 63.4          & 39.8          & 62.0          & 74.7          \\
    Relation-DINO~\cite{hou2024relation}                                   & 57.8 & 76.1 & 62.9       & 41.2 & 62.1 & 74.4 \\
    MDS-DETR (Ours)                        &  
    58.1          &      76.2     & 63.6         & 42.6          &          62.7 & 74.7          \\
    \bottomrule
  \end{tabular}
  }
\end{table}

While $\mathbf{e}_{ij}$ becomes an all-zero vector when $i = j$, it is inherently non-commutative. For example, switching $i$ and $j$ in the input reverses the signs of the $\log(w_i/w_j)$ and $\log(h_i/h_j)$ terms, resulting in a significantly different vector. Through our analysis, we observe that the scale-asymmetry of $\mathbf{e}_{ij}$ significantly degrades performance, particularly for small objects. To ensure the symmetry for $\mathbf{b}_i$ and $\mathbf{b}_j$, we enforce commutativity by defining the relative position bias $R_{ij}$ as:
\begin{equation}
R_{ij} = \frac{1}{2} \left[ \mathrm{MLP}\big(\mathrm{SinEnc}(\mathbf{e}_{ij})\big) + \mathrm{MLP}\big(\mathrm{SinEnc}(\mathbf{e}_{ji})\big) \right]
\end{equation}
where $\mathrm{SinEnc}(\cdot)$ denotes the standard sinusoidal positional encoding function for Transformers~\cite{vaswani2017attention}, MLP refers to multi-layer perceptron. Unlike prior works~\cite{hu2018relation, hou2024relation}, we do not apply a ReLU activation to clamp $R_{ij}$. Allowing $R_{ij}$ to take negative values enables the model to penalize attention between distant queries, thereby promoting more effective duplicate suppression among spatially closer queries.

Given the sorted query set \(\hat{\textbf{q}}^{(l-1)}_i\), we additionally append \(n_t\) true positive tokens to the set. The masked self-attention score is computed via a scaled dot-product with learnable projections \(W_Q\) and \(W_K\), along with a symmetric relative position bias \(\hat{R}_{ij}\). Our masked attention weight \(a_{ij}\) is given by:
\begin{equation}
\begin{aligned}
A_{ij} &= \frac{(W_Q \hat{\mathbf{q}}^{(l-1)}_i)^\top (W_K \hat{\mathbf{q}}^{(l-1)}_j)}{\sqrt{d}} + \hat{R}_{ij}, \\
a_{ij} &= \frac{\hat{M}_{ij} \cdot \exp(A_{ij})}{\sum_{j'=1}^{n + n_t} \hat{M}_{ij'} \cdot \exp(A_{ij'})}
\end{aligned}
\end{equation}

For all token positions \(j > n\), the mask value is set to \(\hat{M}_{ij} = 1\), ensuring they are always visible. In addition, \(\hat{R}_{ij}\) for token positions is implemented as an extra learnable scalar parameter since the true positive token is not spatially grounded. After the MDS, we reverse the sorting operation to restore the original query order.

\section{Experiments}
\subsection{Experiment Settings and Implementation Details}
Building upon prior work, we evaluate our method on the COCO 2017 dataset~\cite{mscoco} using standard AP metrics: mAP, AP$_{50}$, AP$_{75}$, AP$_S$, AP$_M$, and AP$_L$. All experiments were conducted using an ImageNet~\cite{imagenet}-pretrained ResNet-50 backbone~\cite{he2016deep} with a batch size of 16 on 8 NVIDIA RTX 4090 GPUs. We use the AdamW optimizer~\cite{loshchilov2018decoupled} with a learning rate of \(2 \times 10^{-4}\) and weight decay of \(1 \times 10^{-4}\). For the 12-epoch and 24-epoch settings, the learning rate is decayed by a factor of 0.1 at the 11th and 20th epochs, respectively. We adopt a Varifocal loss~\cite{zhang2021varifocalnet} for training.
We follow the same one-to-many matcher used in previous works~\cite{zhang2024mr,zhao2024ms,ouyang2022nms} setting $\alpha = 0.3$, $\tau = 0.4$, and $K = 6$, to ensure fair comparison. 

Following standard DETR settings, MDS-DETR also adopts a classification loss $\mathcal{L}_{\textit{cls}}$ and regression losses $\mathcal{L}_{\textit{gIoU}}$ and $\mathcal{L}_{\textit{l1}}$, based on gIoU and L1, respectively. For classification, we use Varifocal loss~\cite{zhang2021varifocalnet} with commonly used hyperparameters: $\alpha = 0.75$ and $\gamma = 2$. Specifically, MDS-DETR is trained with both one-to-many and one-to-one matching losses. In the one-to-one supervision, the loss weights for $\mathcal{L}_{\textit{cls}}$, $\mathcal{L}_{\textit{gIoU}}$, and $\mathcal{L}_{\textit{l1}}$ are set to 1:5:2, which also applies to the two-stage encoder proposals. In the one-to-many supervision, the loss ratio is set to 0.5:5:2, reducing the weight on classification loss. This adjustment is intended to keep the model focused on one-to-one prediction by down-weighting the influence of one-to-many class probabilities.

\begin{figure}[t]
    \begin{minipage}[t]{0.6\linewidth}
      \centering
      \includegraphics[height=3.4cm]{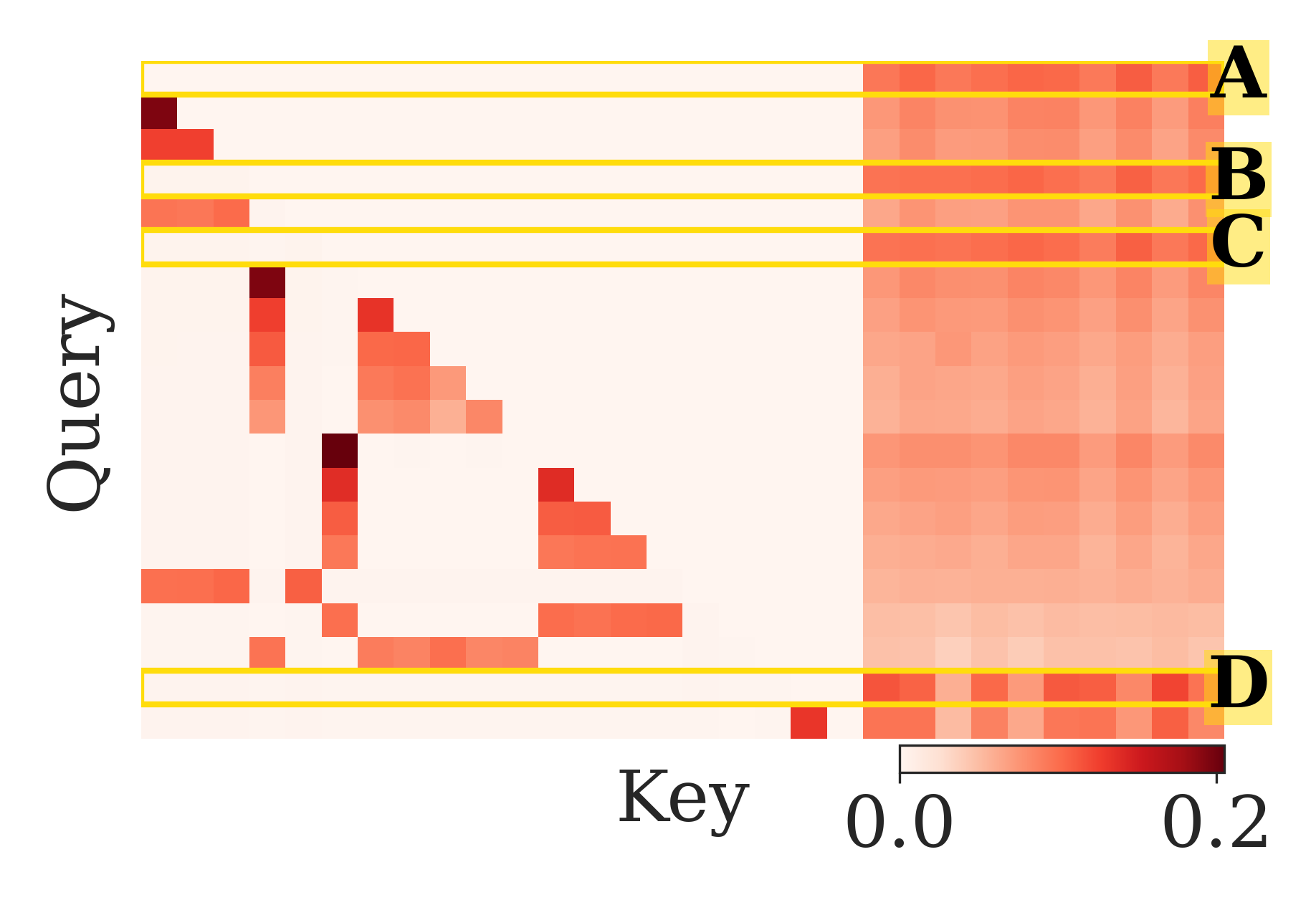}
    \end{minipage}%
    \hfill
    \begin{minipage}[t]{0.27\linewidth}
      \vspace{-95pt}
      \includegraphics[height=2.9cm]{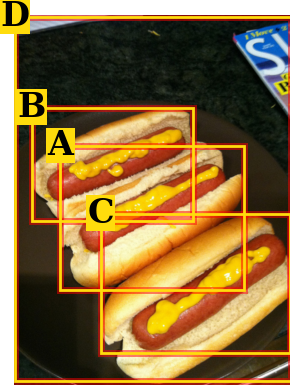}
      \vspace{6pt}
    \end{minipage}
    \vspace{-5pt}
\caption{Visualization of masked self-attention heatmaps for the top 20 one-to-many queries and 10 true positive tokens. True positive queries (A, B, C, D) do not interact with others and are directly guided by the true positive tokens, while the remaining queries are treated as duplicates and suppressed by higher-confidence queries.}
\label{fig:subA}
\end{figure}

\subsection{Comparison with state-of-the-art DETRs}
As shown in \Cref{tab:exp coco2017}, MDS-DETR demonstrates consistent improvements across overall metrics under the 12-epoch schedule with 300 queries. Under the same 12-epoch schedule with 900 queries, MDS-DETR outperforms existing state-of-the-art methods. Compared to the DETR variants with the same one-to-many matching strategies, MDS-DETR achieves improvements of +1.1, +1.1, and +0.4 mAP over MS-DETR, DAC-DETR, and MR.DETR, respectively. When combined with hybrid matching ~\cite{jia2023detrs}, Hybrid-MDS-DETR also yields a marginal performance gain. Furthermore, under the 24-epoch training schedule, MDS-DETR outperforms MR.DETR and Relation-DETR, both combined with DINO, by a margin of +0.6 mAP and +0.2 mAP. We emphasize that all of our results are achieved without training any auxiliary decoders or additional queries. For instance, Relation-DETR and MR.DETR require 1,500 and 1,800 auxiliary queries during training, respectively, and additionally adopt around 200 contrastive denoising queries. 

We report extended results using the Swin-L~\cite{liu2021swin} backbone in \Cref{tab:swin}. MDS-DETR outperforms Relation-DINO~\cite{hou2024relation} by +0.3 mAP and +1.4 AP$_S$ using the Swin-L backbone. While MDS-DETR performs on par with MR.DINO~\cite{zhang2024mr} in overall mAP, we observe that the performance gap varies significantly depending on object size. Notably, MDS-DETR outperforms the others with a significant margin on small objects, surpassing MR.DINO by +1.4 AP and other metrics remain comparable. Considering that small object detection is generally regarded as more challenging, we argue that our model exhibits better generalizability. MDS-DETR requires no additional components or hyperparameters, whereas Relation-DINO and MR.DINO rely on auxiliary decoders (as detailed in Table 1) and introduce extra hyperparameters from IA-BCE loss~\cite{cai2023align} and DINO~\cite{zhang2022dino}.

\begin{figure}[t]
\centering
  \begin{subfigure}[t]{0.55\linewidth}
    \centering
    \includegraphics[width=\linewidth]{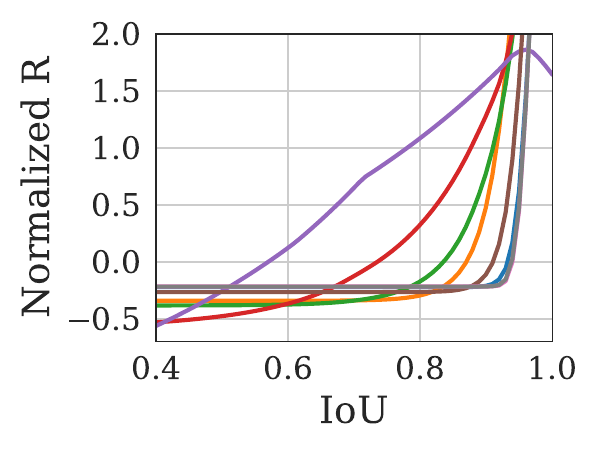}
    \caption{Headwise relative position bias}
    \label{fig:subB}
  \end{subfigure}
  \hfill
  \begin{subfigure}[t]{0.41\linewidth}
    \centering
    \includegraphics[width=\linewidth]{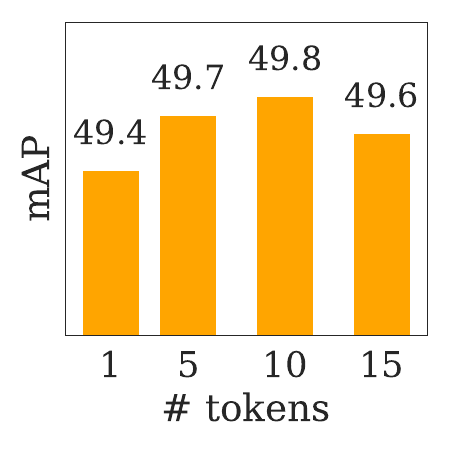}
    \caption{Ablation on tokens}
    \label{fig:ntoken}
  \end{subfigure}
  \caption{(a) Visualization of the relative position bias as a function of the IoU between query pairs. These diverse patterns imply that the model learns a mixture of suppression strategies across attention heads, combining soft and hard thresholding behavior depending on the context. (b) Effect of the number of true positive tokens}
  \label{fig:main}
\end{figure}

\subsection{Qualitative analysis of masked duplicate suppressor}
\Cref{fig:subA} illustrates how MDS effectively maps one-to-many queries into one-to-one predictions. We visualize the normalized attention map from MDS for the top 20 one-to-many queries, where the last 10 keys correspond to true positive tokens. As intended, true positive queries primarily attend to the true positive tokens rather than to other queries within the attention scope. The remaining queries are suppressed by attending to higher-confidence queries that share the same context. Despite substantial overlaps between objects in the scene, the detection results show that our model performs reliable prediction without oversuppression. Unlike the one-to-many instruction token used in \cite{zhang2024mr}, our true positive token operates in a more active and tractable manner, exhibiting a differentiated interaction pattern between true positives and duplicates.

In \Cref{fig:subB}, we plot the normalized average relative position bias \( R_{ij} \) as a function of the IoU between query pairs for various attention heads. The increasing bias with increasing box overlap suggests that the model learns to assign stronger suppression signals to closely overlapping queries. Moreover, the diversity in slope behaviors across attention heads indicates that MDS adaptively handles varying levels of query overlap. Our model learns to suppress duplicates solely through one-to-many and one-to-one supervision signals and attention masking.

We emphasize that the masked self-attention in MDS is both intuitive and explainable.
In previous DETR studies, even it was evident that self-attention played a key role in duplicate suppression by modeling interactions between queries, its attention maps were often intractable and treated as a black box. In contrast, our masked self-attention focuses purely on the similarity between object queries, providing a clear and interpretable mechanism for duplicate suppression.

\definecolor{Gray}{gray}{0.87} 
\begin{table}[t]
\centering
  \caption{Comparison of GPU memory usage and elapsed training time per epoch. All results are measured with a ResNet-50 backbone and 300 queries.}

  \label{tab:ablation4}
  {
  \resizebox{\columnwidth}{!}{
  \begin{tabular}{@{}lccc@{}}
    \toprule
     Model  & Memory (MB)& Duration (min/ep)   & mAP             \\ 
      \midrule
     Def-DETR++ & 7534 & 40 & 47.0 \\
     $\mathcal{H}$-DETR&  11200 & 57 (+17) & 48.7 \\
     MS-DETR & 8326 & 49 (+9) & 48.8 \\
     MR.DETR& 8920 & 53 (+13) & 49.5 \\
     \rowcolor{Gray}
     \textbf{MDS-DETR} & 8252 & 42 (+2) & 49.8 \\
    \bottomrule
  \end{tabular}
  }
  }
\end{table}


\begin{table}[t]
\centering
  \caption{Comparison of Masked Duplicate Suppressor with alternative methods. Ours with 'diag unmasked' refers to MDS with commonly used masked self-attention without diagonal masking.}
  \label{tab:ablation1}
  \resizebox{\columnwidth}{!}{
  \begin{tabular}{@{}llccccc@{}}
    \toprule
    Method                                        & mAP & AP$_{50}$         & AP$_{75}$            \\ 
    \midrule
    Self-attention (baseline)             & 47.7 & 65.2 & 51.4 \\    
    Query-Rank (Rank-DETR)             & 48.5 (+0.8) & 65.9 & 52.8 \\
    MSelf-Attention (Ease-DETR)             & 49.0 (+1.3) & 66.3 & 52.9 \\  
    Ours (diag unmasked)    & 48.0 (+0.3) & 64.8 & 52.3 \\ 
    Ours             & 49.8 (+2.1) & 67.1 & 54.2  \\  
    
    \bottomrule
  \end{tabular}
  }
  \vspace{-5pt}
\end{table}

\begin{table}[t]
  \centering
  \begin{minipage}[t]{0.47\columnwidth}
    \centering
    \caption{Comparison of different positional priors}
    \label{tab:ablation_embedding_bias}
    \resizebox{\linewidth}{!}{%
      \begin{tabular}{@{}ccr@{}}
        \toprule
        Abs P.E & Rel bias & mAP \\
        \midrule 
            &               & 49.3 \\
        \checkmark &         & 49.4 \\
        \checkmark & \checkmark & 49.7 \\
            & \checkmark     & 49.8 \\
        \bottomrule
      \end{tabular}
    }
  \end{minipage}
  \hfill
  \begin{minipage}[t]{0.49\columnwidth}
    \centering
    \caption{Effect of experimental techniques}
    \label{tab:ablation_techniques}
    \resizebox{\linewidth}{!}{%
      \begin{tabular}{@{}lccc@{}}
        \toprule
        DP0 & LFT & MQS & mAP \\
        \midrule
            &      &      & 49.0 \\
        \checkmark &      &      & 49.8 \\
        \checkmark & \checkmark &      & 49.8 \\
        \checkmark & \checkmark & \checkmark & 49.3 \\
        \bottomrule
      \end{tabular}
    }
  \end{minipage}
\end{table}


\subsection{Computational complexity}
\Cref{tab:ablation4} compares the memory usage and training time of MDS-DETR against the Deformable-DETR++ baseline and other DETR variants. Hybrid-DETR incurs 48\% more memory consumption and over 40\% additional training time per epoch compared to the baseline. Since Rank-DETR and Relation-DETR are both built on Hybrid-DETR, they require even higher training costs. MS-DETR avoids auxiliary decoders or queries, but suffers a 22\% slowdown due to enriched supervision on intermediate outputs. MR.DETR adopts a multi-route training scheme with two auxiliary decoders, resulting in a 33\% increase in training time. Although these methods deliver significant performance improvements, they come at the cost of substantial computational overhead.

In contrast, our MDS-DETR requires only 5\% more training time and 9\% more memory usage, while still outperforming previous state-of-the-art DETRs. Notably, MDS-DETR is even faster than MS-DETR, which does not rely on auxiliary decoders, so clearly demonstrating the efficiency of our design.

\subsection{Ablation studies}
\paragraph{Various designs for masked duplicate suppressor.}

To validate the effectiveness of our proposed MDS, we conduct a series of ablation studies on alternative designs. As shown in \Cref{tab:ablation1}, we use a model with conventional self-attention as our baseline, which achieves a relatively low performance of 47.7 mAP. The Query Rank Layer from Rank-DETR~\cite{pu2024rank} and the MSelf-attention module from Ease-DETR~\cite{gao2024ease} yield improvements of +0.8 and +1.3 mAP over the baseline, respectively. In comparison, our MDS achieves a larger gain of +2.0 mAP, demonstrating its effectiveness.

We further analyze the impact of our strict masking strategy by evaluating a variant called diagonal-unmasked MDS, where queries are allowed to attend to themselves. This design leads to a notable drop of 1.7 mAP compared to the original MDS, confirming that preventing self-attention is critical for effective duplicate suppression. These results suggest that our MDS, equipped with strict masking, is a specialized and effective module for this purpose. 

We analyze the performance variation with respect to the number of true positive tokens in \cref{fig:ntoken}. The best performance is achieved with 10 tokens, while using 5 or 15 tokens still outperforms previous DETRs, demonstrating the robustness of MDS. With only a single true positive token, the expressiveness becomes insufficient to instruct true positives, leading to performance degradation.

In \cref{tab:ablation_embedding_bias}, we compare the performance of MDS under different positional priors, including the conventional absolute positional embedding (Abs P.E) and our symmetric relative position bias (Rel bias).
Interestingly, even without any explicit positional information, the performance drop is marginal, suggesting that our confidence-based attention masking is effective independently of positional biases.
Furthermore, we observe that the relative position bias is more effective than the absolute positional embedding for duplicate suppression. However, when both types of positional information are combined, the performance slightly degrades.

\begin{table*}[t]
  \caption{Layerwise performances of MDS-DETR with and without Non-Maximum Suppression (NMS). One-to-many (O2M) and one-to-one (O2O) refers to the matching method for decoder supervision. }
  \label{tab:sup5}
  \centering
  \resizebox{0.8\textwidth}{!}{
  \begin{tabular}{@{}ccccccccc@{}}
    \toprule
    Layer \hspace{10pt} & Type & NMS Post-process \hspace{10pt} & mAP & AP$_{50}$ & AP$_{75}$ & AP$_S$ & AP$_M$ & AP$_L$ \\
    \cmidrule(r){1-3} \cmidrule(l){4-9}
    \multirow{2}{*}{1} & \multirow{2}{*}{One-to-many} &  & 15.1 & 19.0 & 16.4 & 14.8 & 22.4 & 21.9 \\
                       &                     & \checkmark & 47.5 & 65.3 & 51.5 & 30.5 & 51.3 & 61.5 \\
    \cmidrule(r){1-3} \cmidrule(l){4-9}

    \multirow{2}{*}{2} & \multirow{2}{*}{One-to-many} &  & 13.3 & 16.8 & 14.3 & 14.3 & 20.5 & 20.4 \\
                       &                     & \checkmark & 48.8 & 66.3 & 52.6 & 31.6 & 52.4 & 63.9 \\
   \cmidrule(r){1-3} \cmidrule(l){4-9}

    \multirow{2}{*}{3} & \multirow{2}{*}{One-to-many} &  & 12.4 & 15.6 & 13.3 & 13.9 & 19.5 & 18.5 \\
                       &                     & \checkmark & 49.3 & 66.9 & 53.2 & 31.9 & 53.0 & 64.3 \\
     \cmidrule(r){1-3} \cmidrule(l){4-9}

    \multirow{2}{*}{4} & \multirow{2}{*}{One-to-many} &  & 11.7 & 14.8 & 12.7 & 13.7 & 18.8 & 17.2 \\
                       &                     & \checkmark & 49.6 & 67.3 & 53.4 & 31.9 & 53.2 & 64.5 \\
     \midrule

    \multirow{2}{*}{5} & \multirow{2}{*}{One-to-many} &  & 11.4 & 14.6 & 12.3 & 13.4 & 18.9 & 17.1 \\
                       &                     & \checkmark & 49.7 & 67.5 & 53.5 & 32.4 & 53.4 & 64.7 \\
     \cmidrule(r){1-3} \cmidrule(l){4-9}

    \multirow{2}{*}{6} & \multirow{2}{*}{One-to-one} &  & 49.8 & 67.1 & 54.2 & 32.2 & 53.6 & 65.2 \\
                       &                     & \checkmark & 49.6 & 67.4 & 53.0 & 32.3 & 53.0 & 64.8 \\
    \bottomrule
  \end{tabular}
  }
\end{table*}

\paragraph{Experimental techniques for DETRs.}
We further evaluate the impact of three common techniques, removing dropout (DP0), Look Forward Twice (LFT), and Mixed Query Selection (MQS) on the performance of MDS-DETR, as shown in \cref{tab:ablation_techniques}. Consistent with prior works, removing dropout (DP0) yields performance improvements. On the other hand, LFT does not lead to noticeable performance gains. This may be attributed to the nature of one-to-many matching, where the distribution of matched queries remains largely unchanged across decoder layers. 

Interestingly, although MQS typically brings performance gains of over +0.5 mAP in many prior methods~\cite{zhang2022dino}, it degrades our model's performance by -0.4 AP. As noted in Ease-DETR~\cite{gao2024ease}, the learnable embeddings introduced by MQS can be interpreted as a form of rank embedding when applied to two-stage Deformable-DETR that extract top-k proposals from the encoder. However, our MDS also operates explicitly based on the confidence rank of one-to-many queries. We conjecture that this inherent conflict between the MQS rank and our dynamic confidence-based suppression in MDS leads to the observed performance degradation.

\begin{table}[t]
  \caption{Ablation studies on number of one-to-many supervised decoder layers. We vary the number of one-to-many (O2M) and one-to-one (O2O) supervised decoder layers, denoted as \# O2M / \# O2O (e.g., 5/1 or 5/2), to study their impact on performance.}
  \vspace{4pt}
  \label{tab:sup3}
  \resizebox{\columnwidth}{!}{
  \centering
  {
  \begin{tabular}{@{}cccccccc@{}}
    \toprule
     \# O2M / \# O2O   &  1 / 1 & 2 / 1 & 3 / 1 & 4 / 1 & 5 / 1 & 5 / 2 & 5 / 3 \\
     \cmidrule(r){1-1} \cmidrule(l){2-8}
     mAP & 46.4 & 48.1 & 48.9 & 49.4 & 49.8 & 49.7 & 49.7 \\
    \bottomrule
  \end{tabular}
  }
  }
\end{table}

\paragraph{Analysis on number of decoder layers}
\Cref{tab:sup3} presents the effect of varying the number of one-to-many (O2M) and one-to-one (O2O) supervised decoder layers on detection performance. Notably, our MDS-DETR outperforms the 6-layer Deformable-DETR++ baseline using only three decoder layers (2 O2M / 1 O2O). Furthermore, with just four decoder layers (3 O2M / 1 O2O), MDS-DETR surpasses MS-DETR~\cite{zhao2024ms} by +0.3 mAP. We attribute this scalability to the effectiveness of the Masked Duplicate Suppressor (MDS), which enables our O2O layer to perform duplicate suppression in a single stage. In contrast, previous DETR variants suffer a significant drop in precision when the number of decoder layers is reduced. This indicates that conventional decoder layers struggle to suppress duplicates, making multi-stage O2O supervision essential to progressively filter false positives. In our design, however, MDS allows for a much more flexible decoder configuration. 

We also observe that stacking multiple layers with MDS with one-to-one supervision does not lead to further improvements, even it results in a slight performance drop. Since the predictions from queries that have passed through a single MDS are already close to duplicate-free one-to-one outputs, applying additional MDS blocks appears to cause overconfident or oversuppressed predictions.


\definecolor{Gray}{gray}{0.87}
\begin{table}[t]
\centering
\caption{Comparison on instance segmentation task on COCO \texttt{val2017}.}
\label{tab:sup2}
\resizebox{\columnwidth}{!}{
\begin{tabular}{@{}lcccc@{}}
\toprule
\multirow{2}{*}{ Model} & \multicolumn{2}{c}{Epoch 12} & \multicolumn{2}{c}{Epoch 24} \\
\cmidrule(lr){2-3} \cmidrule(lr){4-5}
 & Mask mAP & Box mAP & Mask mAP & Box mAP \\
\midrule
Def-DETR & 32.4 & 46.5 & 35.1 & 48.6 \\
MR.DETR         & 36.0 & 49.5 & 37.6 & 50.3 \\
\rowcolor{Gray}
\textbf{MDS-DETR}          & \textbf{36.4} (+4.0) & \textbf{49.5} (+3.0) & \textbf{38.1} (+3.0) & \textbf{50.6} (+2.0) \\
\bottomrule
\end{tabular}
}
\end{table}
\begin{figure*}[t]
    \centering
    {\includegraphics[width=0.3\textwidth, trim=0pt 200pt 0 50pt, clip]{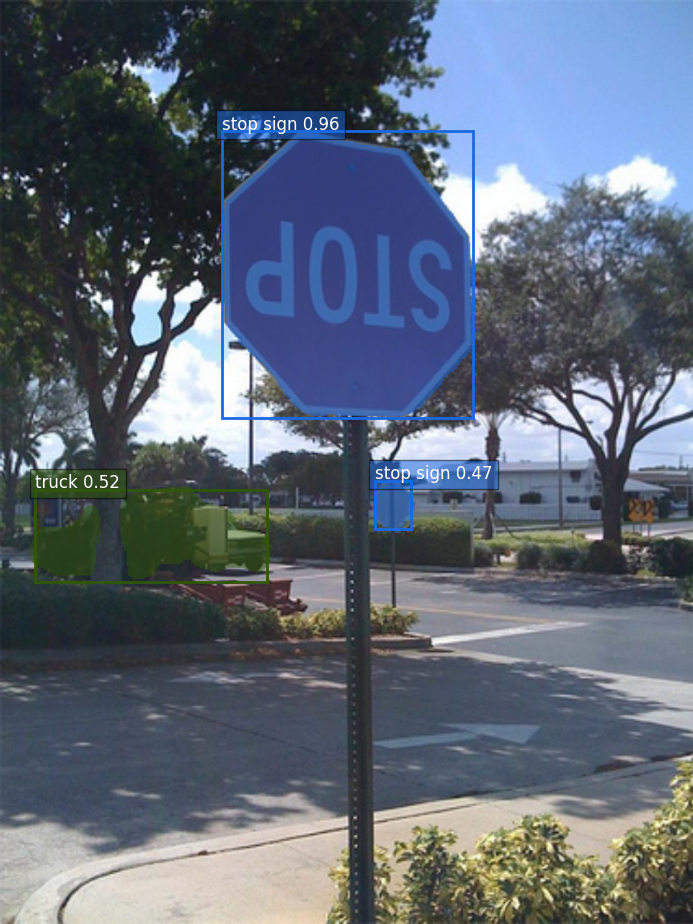}}\hfill
    {\includegraphics[width=0.3\textwidth, trim=15pt 120pt 15pt 15pt, clip]{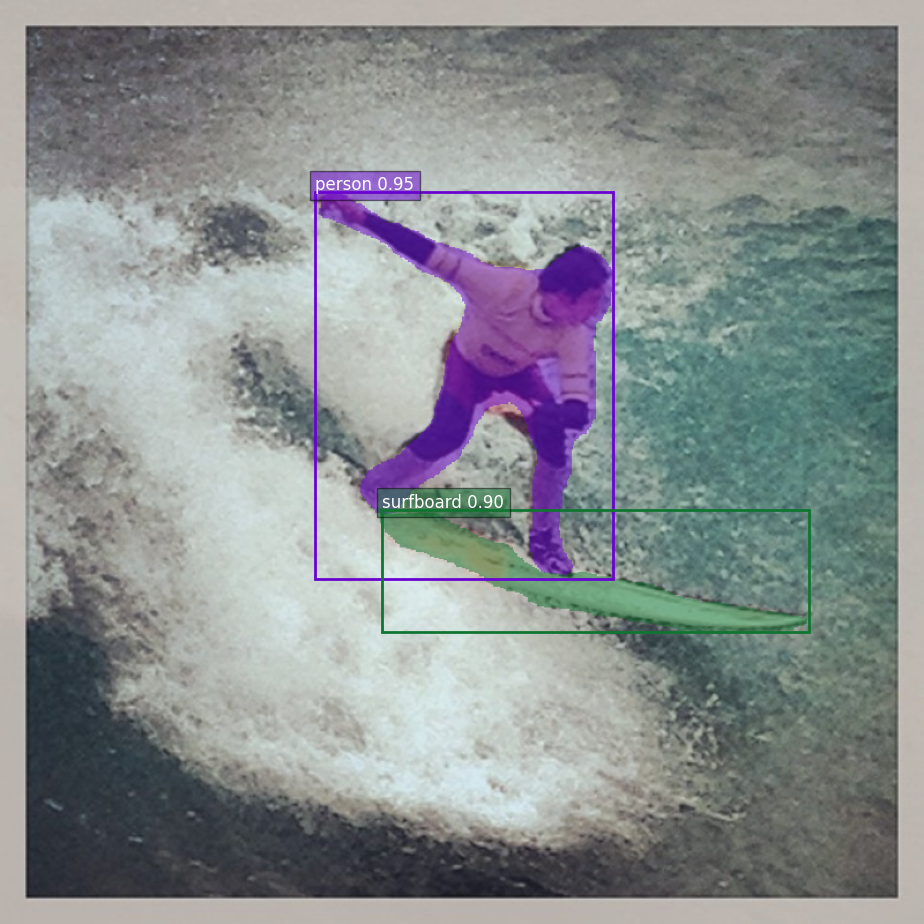}}\hfill
    {\includegraphics[width=0.3\textwidth, trim=105pt 0 30pt 0, clip]{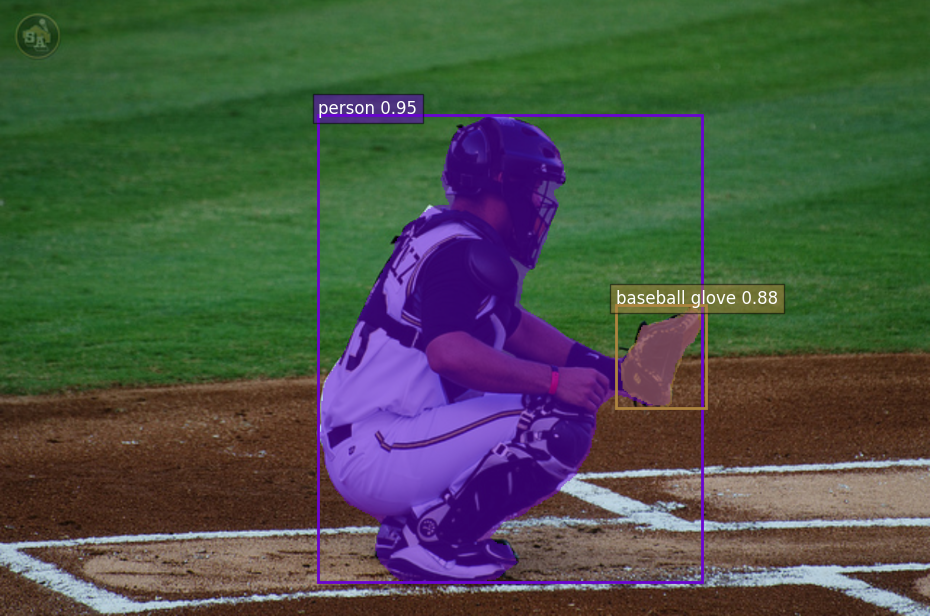}}\\
    \vspace{5pt}
    {\includegraphics[width=0.3\textwidth]{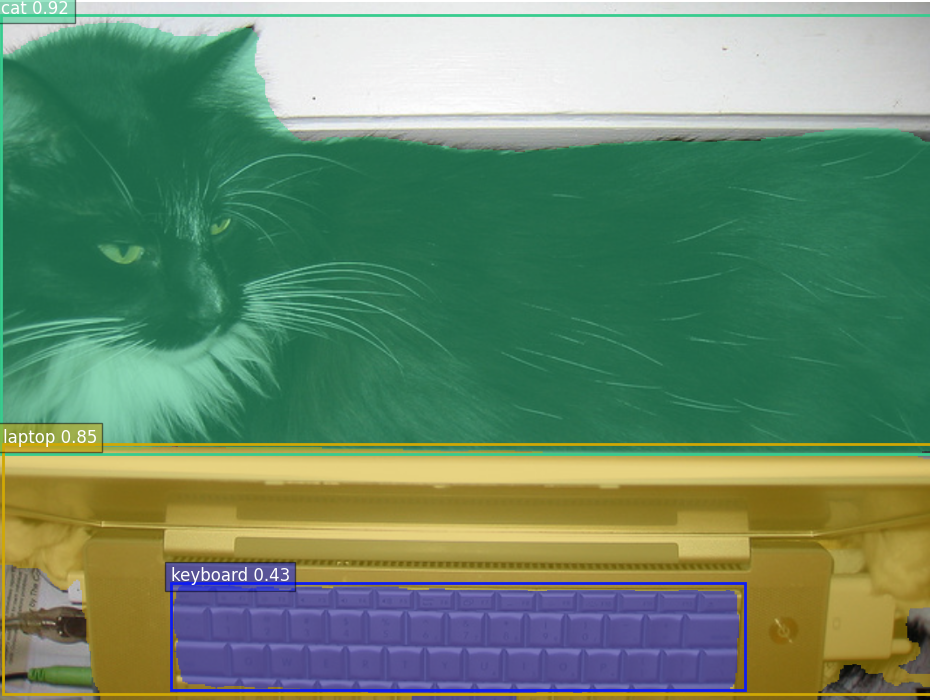}}\hfill
    {\includegraphics[width=0.3\textwidth]{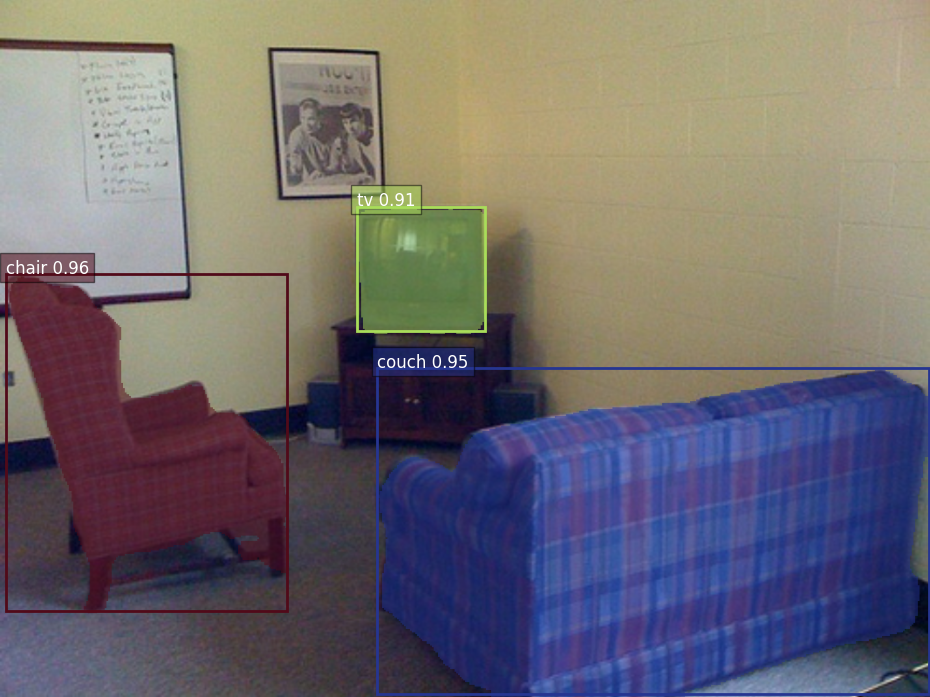}}\hfill
    {\includegraphics[width=0.3\textwidth]{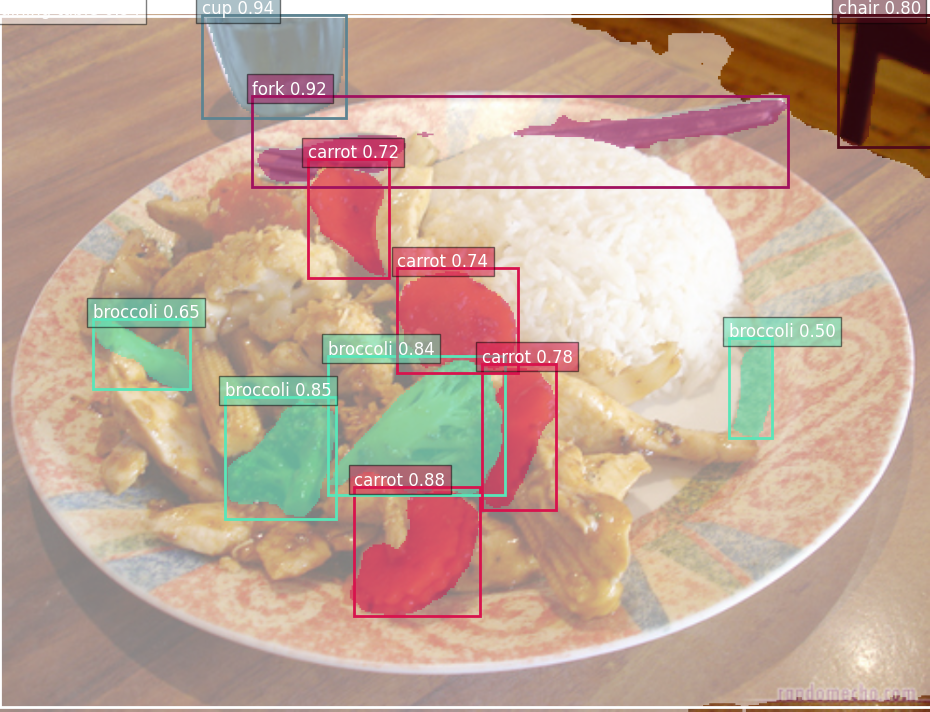}}
    \caption{Qualitative results on COCO \texttt{val2017} instance segmentation with a ResNet-50 backbone.}
    \label{fig:subfig_2x3}
\end{figure*}

\paragraph{Effect of Non-Maximum Suppression on MDS-DETR}

\Cref{tab:sup5} evaluates the layer-wise intermediate outputs of MDS-DETR by applying Non-Maximum Suppression(NMS) postprocessing at each decoder layer. As expected, applying NMS to the predictions from one-to-many layers yields higher performance at deeper layers, indicating that more refined candidates are generated as the decoder progresses. Interestingly, when NMS is not applied, mAP tends to decrease across layers. This suggests that one-to-many layers increasingly produce more candidates with high-quality, so they are not directly optimized for the duplicate-free prediction.

While DETR variants with one-to-many supervision preserve the NMS-free end-to-end property in their main prediction branch, it is often observed that combining the one-to-many predictions with NMS yields better results than the final output~\cite{jia2023detrs,ouyang2022nms,zhang2024mr}. Notably, the final output of MDS-DETR outperforms any intermediate one-to-many layer even when NMS is applied. Furthermore, it does not benefit from NMS, even its performance slightly degrades. Unlike NMS, our MDS module does not rely on hand-crafted post-processing. Instead, it naturally learns duplicate suppression from one-to-one matching supervision in an end-to-end manner. This allows MDS-DETR to achieve higher precision, particularly on strict metrics such as AP$_{75}$ and AP$_S$.

\subsection{Instance Segmentation}
We evaluate an extended version of our MDS-DETR for instance segmentation by attaching an additional mask prediction head, while keeping the original one-to-one or one-to-many matching strategy unchanged. As shown in \cref{tab:sup2}, our MDS-DETR consistently outperforms MR.DETR under both the 12-epoch and 24-epoch training schedules. Notably, while MR.DETR applies instance mask supervision across all three decoders with extra complexity, our model achieves large gains in both box mAP and mask mAP while maintaining nearly the same training cost as the Deformable-DETR baseline. This demonstrates the effectiveness of our suppression-aware design, even when extended to dense prediction tasks like instance segmentation.

\section{Conclusion}
To better exploit one-to-many supervision in DETR-based detectors, we explored a strategy that blends both one-to-many and one-to-one supervision within a single decoder. At the core of this approach is the proposed Masked Duplicate Suppressor (MDS), which effectively bridges the gap between one-to-many and one-to-one predictions through an asymmetric, confidence-based masking mechanism. Our experiments demonstrate that MDS-DETR not only improves detection performance, but also offers a simple yet effective pipeline that eliminates the need for auxiliary decoders and queries, significantly reducing training cost. Despite these advantages, MDS-DETR still has room for improvement. For instance, integrating techniques such as query denoising or mixed query selection into our pipeline is non-trivial due to its structural differences. We leave the extension of MDS-DETR to DINO or other DETR variants as our future work.

\section*{Acknowledgment}
This work was partly supported by Center for Applied Research in Artificial Intelligence (CARAI) grant funded by Defense Acquisition Program Administration (DAPA) and Agency for Defense Development (ADD) (UD230017TD) and a grant of the Korea Health Technology R\&D Project through the Korea Health Industry Development Institute (KHIDI), funded by the Ministry of Health \& Welfare, Republic of Korea (grant number: RS-2022-KH129703).

\bibliographystyle{IEEEtran}
\bibliography{access}

\begin{IEEEbiography}[{\includegraphics[width=1in,height=1.25in,clip,keepaspectratio]{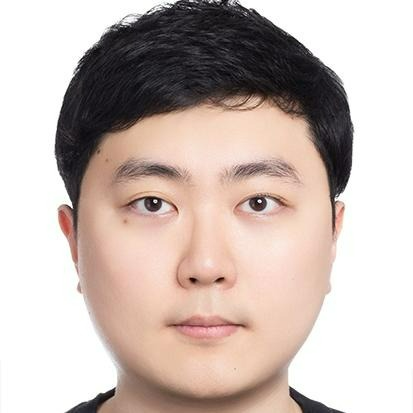}}]{CHANHO LEE} received the B.S., M.S., and Ph.D. degrees in electrical engineering from the Korea Advanced Institute of Science and Technology (KAIST), Daejeon, South Korea, in 2016, 2019, and 2026, respectively. His research interests include computer vision, object detection, image segmentation, AI perception, vision grounding, and vision-language models (VLMs). He has authored publications in top-tier AI conferences, including work on oriented object detection presented at the AAAI Conference on Artificial Intelligence. He is currently a staff engineer on computer vision software with Samsung Research, Seoul, South Korea.
\end{IEEEbiography}

\begin{IEEEbiography}[{\includegraphics[width=1in,height=1.25in,clip,keepaspectratio]{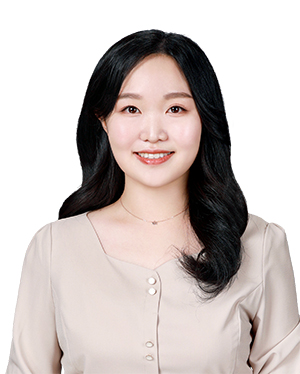}}]{SEUNGHEE KOH} received the B.S. and M.S. degrees in electrical engineering from the Korea Advanced Institute of Science and Technology (KAIST), Daejeon, South Korea, in 2021 and 2023, respectively. She is currently a Ph.D. student in electrical engineering at KAIST, Daejeon, South Korea. Her research interests include medical imaging, especially CT image segmentation, and trustworthy AI.
\end{IEEEbiography}

\begin{IEEEbiography}[{\includegraphics[width=1in,height=1.25in,clip,keepaspectratio]{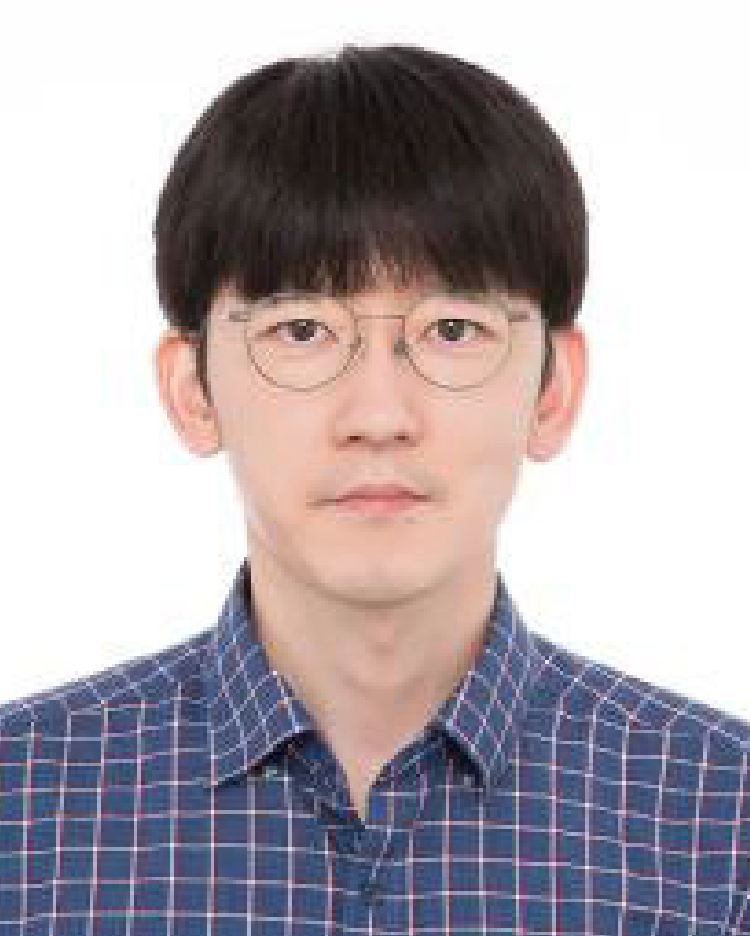}}]{YUNHO JEON} received the B.S. degree in mathematics and the M.S. degree in computer science from Seoul National University, in 2006 and 2008, respectively, and the Ph.D. degree in electrical engineering from Korea Advanced Institute of Science and Technology (KAIST), in 2019. From 2008 to 2019, he was a Senior Researcher with the Agency for Defense Development, South Korea. After receiving the Ph.D. degree, he was with SK Telecom, in 2019, and then at mofl Inc., from 2019 to 2022. He has been an Assistant Professor with the Department of Artificial Intelligence Software, Hanbat National University, South Korea, since 2023. His research interests include image processing, computer vision, time series analysis, and deep learning.
\end{IEEEbiography}

\begin{IEEEbiography}[{\includegraphics[width=1in,height=1.25in,clip,keepaspectratio]{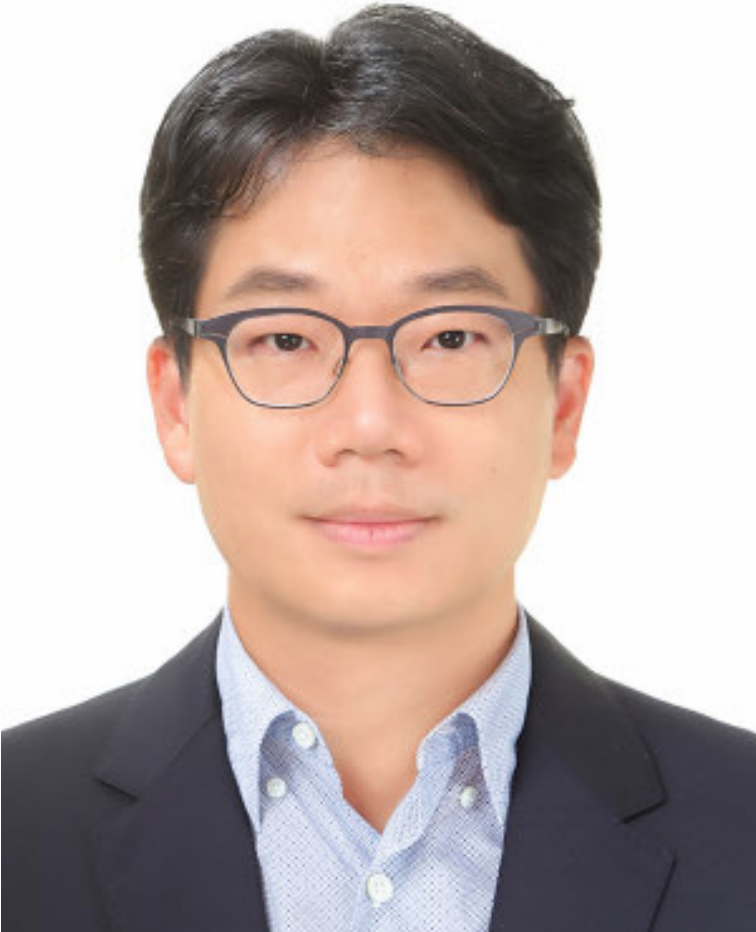}}]{JUNMO KIM} (Member, IEEE) received the B.S. degree from Seoul National University, South Korea, in 1998, and the M.S. and Ph.D. degrees from Massachusetts Institute of Technology, in 2000 and 2005, respectively.
From 2005 to 2009, he was with the Samsung Advanced Institute of Technology, South Korea.
In 2009, he joined Korea Advanced Institute of Science and Technology, as a Faculty Member, where he is currently a Tenured Associate Professor of electrical engineering. His current research interests include image processing, computer vision, statistical signal processing, machine learning, and information theory.
\end{IEEEbiography}

\EOD

\end{document}